\documentclass[a4paper,fleqn]{cas-dc}

\usepackage[numbers,sort&compress]{natbib}
\usepackage[commandnameprefix=always]{changes}
\usepackage{subcaption}



\usepackage{tabularx} % 放在导言区
\usepackage{amsthm}
\usepackage{array}
\usepackage{algorithm}
\usepackage{algpseudocode}
\usepackage{booktabs}
\usepackage{threeparttable}
\usepackage[normalem]{ulem}

\theoremstyle{remark}          % 让正文用直立体
\newtheorem{remark}{Remark}    % 带编号
\hypersetup{
  colorlinks=false,        % 不给文字染色，只用方框
 pdfborder={0 0 1},       % 框线宽度 1pt
            citebordercolor={0 1 0}, % citation 绿色框
            linkbordercolor={1 0 0}, % \ref 等也绿色框
            urlbordercolor={0 1 1}   % URL 绿色框
}

\begin{document}
\let\WriteBookmarks\relax
\def\floatpagepagefraction{1}
\def\textpagefraction{.001}

\shorttitle{HuMam: Humanoid Motion Control via End-to-End Deep Reinforcement Learning with Mamba}

\shortauthors{Yinuo Wang, Yuanyang Qi, Jinzhao Zhou,  and Xiaowen Tao}

\title [mode = title]{HuMam: Humanoid Motion Control via End-to-End Deep Reinforcement Learning with Mamba}                      

% Conceptualization
% Methodology
% Software
% Validation
% Formal analysis
% Investigation
% Resources
% Data Curation
% Writing - Original Draft
% Writing - Review & Editing
% Visualization
% Supervision
% Project administration
% Funding acquisition

\author[1]{Yinuo Wang}[style=chinese]
\ead{allen.wang@woven.toyota}
\credit{Conceptualization, Methodology, Software, Validation, Investigation, Visualization, Writing- Original Draft, Writing- Review \& Editing}

\author[2]{Yuanyang Qi}[style=chinese]
\ead{u3011178@connect.hku.hk}
\credit{Writing- Original Draft, Writing- Review \& Editing}

\author[3]{Jinzhao Zhou}[style=chinese]
\ead{jinzhao.zhou@uts.edu.au}
\credit{Conceptualization, Writing- Original Draft, Writing- Review \& Editing}

\author[4]{Pengxiang Meng}[style=chinese]
\ead{mengpx21@mails.jlu.edu.cn}
\credit{Writing- Review \& Editing}

\author[5]{Xiaowen Tao}[style=chinese]
\cormark[1]
\ead{taox@tcd.ie}
\credit{Conceptualization, Methodology, Formal analysis, Writing- Original Draft, Writing- Review \& Editing, Supervision}

\affiliation[1]{organization={College of Graduate and Professional Studies, Trine University},
    city={Detroit},
    state={Michigan},
    country={USA}}

\affiliation[2]{organization={Department of Civil Engineering, University of Hong Kong},
    city={Hong Kong SAR},
    country={China}}

\affiliation[3]{organization={Faculty of Engineering and Information Technology, University of Technology Sydney},
    city={Sydney},
    country={Australia}}

\affiliation[4]{organization={National Key Laboratory of Automotive Chassis Integration and Bionics, Jilin University},
    city={Changchun},
    country={China}}
    
\affiliation[5]{organization={School of Computer Science and Statistics, Trinity College Dublin},
    city={Dublin},
    % citysep={}, % Uncomment if no comma needed between city and postcode
    country={Ireland}}

\cortext[cor1]{Corresponding author}

\begin{abstract}
End-to-end reinforcement learning (RL) for humanoid locomotion is appealing for its compact perception–action mapping, yet practical policies often suffer from training instability, inefficient feature fusion, and high actuation cost. We present \emph{HuMam}, a state-centric end-to-end RL framework that employs a single-layer Mamba encoder to fuse robot-centric states with oriented footstep targets and a continuous phase clock. The policy outputs joint position targets, tracked by a low-level proportional–derivative controller, and is optimized using proximal policy optimization. A concise six-term reward balances contact quality, swing smoothness, foot placement, posture, and body stability, while implicitly promoting energy efficiency. On the JVRC-1 humanoid in \texttt{mc-mujoco}, HuMam consistently improves learning efficiency, training stability, and overall task performance over a strong feedforward baseline, while reducing power consumption and torque peaks. To our knowledge, this is the first end-to-end humanoid RL controller that adopts Mamba as the fusion backbone, demonstrating tangible gains in efficiency, stability, and control economy. A repository is hosted at \url{https://github.com/allen-legged-robot/humam-rl}.

\end{abstract}

% \includegraphics{figs/grabs.pdf}

\begin{highlights}
\item \textbf{First Mamba-backed humanoid RL controller.}
Introduces Mamba as a lightweight fusion backbone for end-to-end humanoid locomotion.

\item \textbf{Efficient and stable backbone.}
Mamba enables faster learning, better sample efficiency, and lower variance than baseline.

\item \textbf{Energy-aware reward shaping.}
A six-term PPO reward balances stability, accuracy, smooth actuation, and energy use.

\item \textbf{Extensive humanoid validation.}
Consistent gains on JVRC-1 across diverse locomotion tasks in \texttt{mc-mujoco}.
\end{highlights}

% Keywords

\maketitle

\section{Introduction}
\label{sec:first}
Humanoid robots are envisioned as versatile platforms for tasks in human-centric environments, including industrial assistance, healthcare, disaster response, and space exploration \cite{castillo2021digit}. Unlike wheeled or quadrupedal systems, humanoids face heightened challenges: their larger mass, higher center of gravity, and coupled upper–lower body dynamics make locomotion both demanding and safety-critical. Developing reliable locomotion controllers is therefore crucial—not only for advancing embodied intelligence, but also for enabling the safe, efficient, and adaptive deployment of humanoids in real-world scenarios.

Humanoid locomotion demands controllers that are both foresightful and resource-aware: foresightful to coordinate accurate foot placement and whole-body balance, and resource-aware to run reliably under onboard compute and actuation limits \cite{tong2024advancements}. End-to-end reinforcement learning (RL) is attractive because it can discover feedback strategies directly from interaction \cite{singh2022reinforcement}; however, its effectiveness hinges on (i) how heterogeneous inputs are fused and (ii) how training is shaped to avoid trivial or unstable behaviors. Conventional feedforward backbones may under-exploit structure in the observation (e.g., gait phase and oriented footsteps), while heavier sequence models impose unfavorable compute–memory trade-offs \cite{saeedvand2019comprehensive}.

We address these challenges with \emph{HuMam}, a compact, state-centric RL framework for humanoid walking. HuMam employs a single-layer Mamba encoder as the backbone to fuse (i) robot-centric states, including leg joint positions and velocities as well as base attitude and angular velocity, and (ii) external guidance specified by two oriented footsteps with three-dimensional position and heading together with a continuous clock that encodes the gait phase. The policy operates at the position level and is executed through a low-gain proportional–derivative (PD) controller, yielding smooth torques while keeping learning well-conditioned \cite{darvish2023teleoperation}. A concise six-term reward promotes contact quality, swing smoothness, precise stepping, upright posture, nominal height, and upper-body stability—encouraging stable, accurate, and energy-efficient gaits.

\textbf{The main contributions of this paper are summarized as follows:}
\begin{enumerate}
\item \textbf{First Mamba-backed end-to-end humanoid RL controller.} We introduced Mamba as a lightweight fusion backbone for humanoid locomotion and show that a single Mamba layer suffices to deliver strong performance in a purely state-centric setting.
\item \textbf{Backbone effectiveness for efficiency \& stability.} Relative to an identically trained feedforward Baseline, the Mamba encoder yields faster learning, improved sample efficiency, and reduced cross-seed variability, culminating in higher final returns under the same training budget.
\item \textbf{Reward shaping for balanced, energy-saving control.} 
A six-term reward for proximal policy optimization (PPO) jointly regulates foot contact forces and swing velocities, stepping accuracy, body orientation, torso height, and upper-body sway, leading to smoother actuation, reduced torque usage, and lower power consumption.

\item \textbf{Extensive empirical validation.} On the JVRC-1 humanoid across forward, backward, curved, lateral, and standing tasks in \texttt{mc-mujoco}, HuMam consistently improves overall per-step reward, energy metrics, and joint-torque profiles, confirming that Mamba provides an efficient and stable backbone for end-to-end humanoid locomotion.
\end{enumerate}

The rest of the paper is structured as follows. 
Section~\ref{sec:second} surveys reinforcement learning for legged and humanoid locomotion. 
Section~\ref{sec:third} details our methodology, including the MDP formulation, observation and action spaces, the single-layer Mamba encoder backbone, reward design, and PPO training. 
Section~\ref{sec:fourth} presents the experimental setup and results, with analyses of performance, learning efficiency and stability, energy use, and torque profiles. 
Section~\ref{sec:fifth} concludes and outlines directions for future work.

\section{Related Work}
\label{sec:second}

\subsection{Legged Motion Control Based on Reinforcement Learning}

Learning-based locomotion has rapidly progressed from simulation-only proofs of concept to robust, real-world controllers on quadrupeds and humanoids. Pioneering sim-to-real pipelines demonstrated that policies trained purely in simulation with appropriate dynamics randomization can execute agile skills on hardware, enabling precise velocity tracking, fast trotting, and self-righting on ANYmal without hand-crafted gait generators \cite{Hwangbo2019Agile}. Subsequent work showed that a \emph{blind}, purely proprioceptive policy learned via RL can achieve remarkable zero-shot robustness in natural outdoor environments---including rubble, mud, snow, vegetation, and flowing water—by processing streams of joint and base states with a compact neural controller \cite{Lee2020RoughTerrain}. 

A complementary line reduces training wall-clock by exploiting massive parallelism in GPU simulators, making on-policy RL practical at scale. By training thousands of agents in parallel with a curriculum over terrains, policies for flat and moderately rough grounds can be obtained in minutes and transferred to the real robot with minimal tuning \cite{Rudin2022Minutes}. Beyond static policies, \emph{Rapid Motor Adaptation} augments an RL base controller with a learned adaptation module that infers latent environment parameters online, enabling real-time adjustment to novel terrains, payload changes, and wear without explicit system identification \cite{Kumar2021RMA}. 

To achieve stable periodic gaits, works such as \cite{siekmann2021sim} employed reward shaping based on clock signals to synchronize swing and stance phases, while others \cite{li2021reinforcement} leveraged parameterized reference motions from Hybrid Zero Dynamics (HZD) to generate robust walking behaviors. Traversing complex terrain such as stairs has also been demonstrated using purely proprioceptive signals \cite{siekmann2021blind}, enabling “blind” locomotion that remains robust to unexpected contacts and missteps. While these works highlight the resilience of proprioceptive-only controllers, they also underscore the lack of anticipatory planning, as policies must react after contacts occur. This motivates the integration of exteroceptive modalities, such as planned footsteps, to enable more foresightful locomotion.

\subsection{Humanoid Motion Control Based on Reinforcement Learning}

Applying RL to humanoid platforms poses unique challenges due to larger link masses, high-gear transmissions, and reduced backdrivability, which increase safety risks and make contact-rich exploration more problematic than in quadrupeds or lightweight bipeds. Early progress has therefore been made primarily in high-fidelity simulation. \cite{Yang2020WholeBody} demonstrated Valkyrie gait learning with demonstration-augmented rewards and tracking costs that bias exploration toward humanlike bipedal walking while still allowing RL to discover stabilizing strategies without hand-crafted gait schedules. Similarly, \emph{DeepWalk} achieved a single omnidirectional walking policy through curriculum scheduling of commanded velocities and contact-aware rewards, producing robust gaits in a simulated humanoid \cite{Rodriguez2021DeepWalk}.

A second line of work adopts hierarchical designs that combine learned high-level policies with model-based low-level regulators. On the Digit biped, Castillo et al.\ proposed a cascade structure in which an RL policy outputs task-space setpoints tracked by an operational-space controller, enabling disturbance rejection and transfer to hardware without extensive dynamics randomization \cite{castillo2021robust}. Complementary curriculum-based studies have highlighted the difficulty of precise stepping tasks. The ALLSTEPS framework introduced a curriculum that gradually increases stepping precision and spacing, yielding robust stepping policies for humanoid and Cassie-like robots \cite{Xie2020ALLSTEPS}. More recently, Duan et al.\ advanced Cassie controllers to closed-loop stepping across challenging stone patterns by mapping footstep commands directly to joint actions, aided by a learned feasibility model and real-time perception \cite{Duan2022SteppingStones}.

To the best of our knowledge, this is the first work to introduce Mamba for end-to-end humanoid locomotion with RL.

\section{Methodology}
\label{sec:third}

\subsection{Overview}
\begin{figure*}
\centering
\includegraphics[clip, trim=100pt 20pt 20pt 10pt, width=1.0\linewidth]{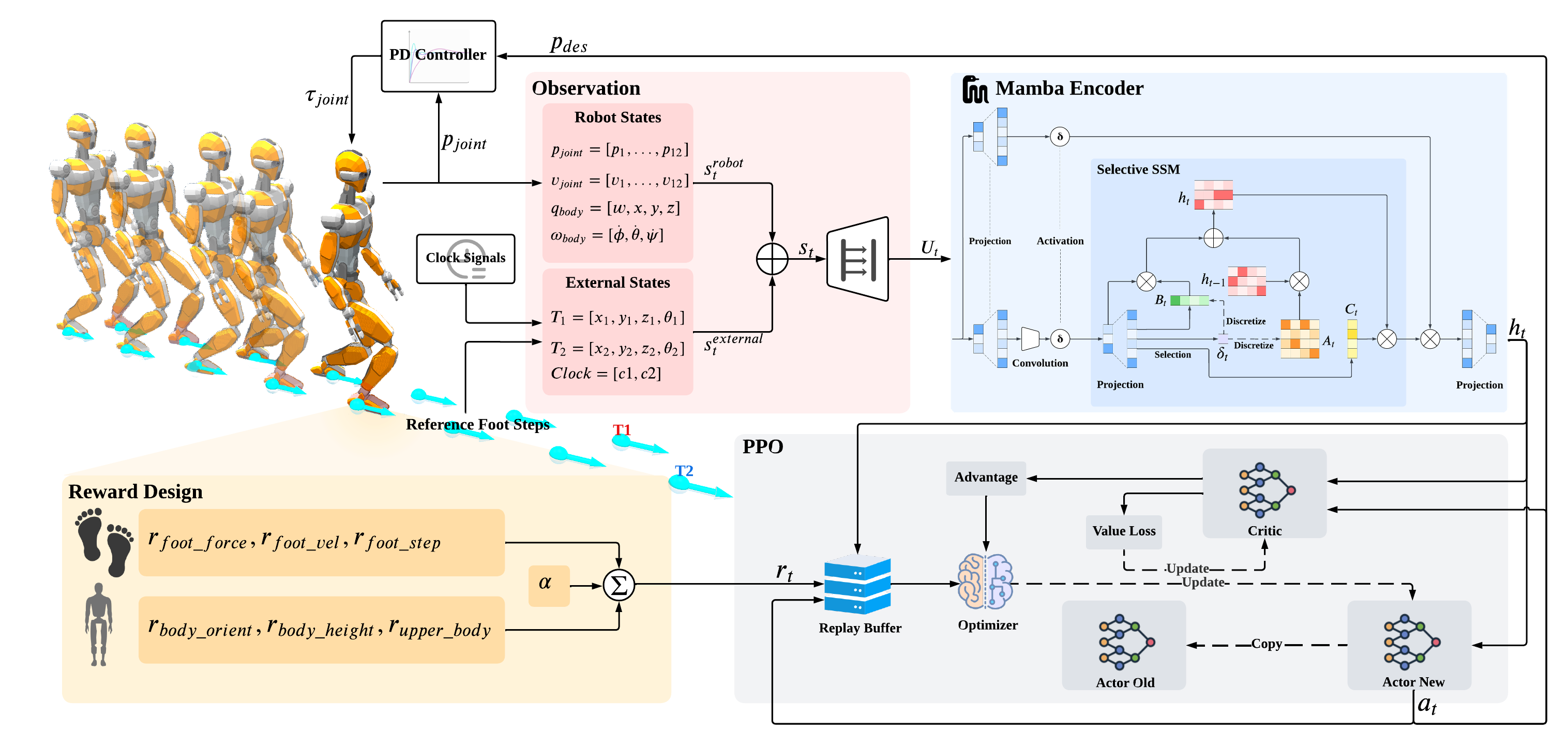}
\caption{Overall architecture of the proposed humanoid locomotion framework. At each time step, robot-centric and external states are collected as observations and projected into a latent embedding. A single-layer Mamba encoder processes these features to produce compact representations for the policy and value heads, which are optimized using PPO. A hierarchical control structure is adopted, where the high-level policy outputs desired joint positions and a low-gain PD controller converts them into executable joint torques. The reward design combines foot-level and body-level objectives to encourage stable and natural gaits.}
\label{fig:overall_archi}
\end{figure*}

We proposed an end-to-end reinforcement learning framework for humanoid locomotion that employs a Mamba encoder as a compact fusion backbone for policy learning. The overall pipeline is illustrated in Fig.~\ref{fig:overall_archi}. At each time step, the agent receives a rich set of observations, consisting of robot-centric states (joint positions, velocities, body orientation, and angular velocity) and external states (target footsteps, root heading, and clock signals). These heterogeneous features are projected into a latent embedding and processed by a single-layer Mamba encoder, which introduces lightweight state-space dynamics to enhance feature mixing while preserving computational efficiency. Unlike recurrent or attention-based designs, this formulation relies solely on the current state, avoiding temporal accumulation yet still capturing structured dependencies across input modalities.

The encoded representation is mapped to policy and value heads that are jointly optimized with PPO. A hierarchical control scheme is adopted, where the high-level policy outputs desired joint positions at 40Hz, while a low-gain PD controller operating at 1000Hz converts them into executable joint torques. The policy naturally incorporates the PD tracking behavior into its predictions, effectively leveraging residual tracking errors as stabilizing interaction forces.

Reward shaping further guides the emergence of robust gaits. Foot-level terms penalize excessive contact forces, encourage smooth swing velocities, and enforce accurate foot placement, while body-level terms promote upright orientation, sufficient body height, and upper-body stability. These objectives are weighted into a scalar reward that balances locomotion stability and efficiency. Together, this design enables the learned controller to synthesize natural and stable humanoid walking behaviors without relying on explicit temporal history.

\subsection{Problem Formulation}

1) \textbf{Markov Decision Process:}  
We model humanoid motion control as a Markov Decision Process (MDP) \cite{singh2022reinforcement}, 
denoted by $\mathcal{M} = (\mathcal{S}, \mathcal{A}, \mathcal{P}, r, \gamma)$. 
Here, $\mathcal{S}$ represents the state space, $\mathcal{A}$ the action space, 
$\mathcal{P}(s_{t+1} \mid s_t, a_t)$ the transition dynamics governed by the physics simulator, 
and $r(s_t,a_t)$ the immediate reward. 
The objective is to optimize a policy $\pi_\theta(a_t \mid s_t)$ that maximizes the cumulative discounted reward:
\begin{equation}
J(\pi) = \mathbb{E}_{\pi}\!\left[\sum_{t=0}^{T-1} \gamma^t r(s_t,a_t)\right],
\end{equation}
where $\gamma \in (0,1)$ denotes the discount factor.

2) \textbf{Observation space:}  
At each time step $t$, the agent receives an observation consisting of robot-centric states and external states:
\begin{equation}
s_t = \Big\{\, s^{\text{robot}}_t,\; s^{\text{external}}_t \,\Big\}.
\end{equation}
The robot state $s^{\text{robot}}_t$ includes joint positions $p_{\text{joint}} \in \mathbb{R}^{12}$ and joint velocities 
$v_{\text{joint}} \in \mathbb{R}^{12}$ of all actuated leg joints, as well as the root orientation 
$\phi_{\text{body}} = [\phi, \theta]$ (roll and pitch) and angular velocity 
$\omega_{\text{body}} = [\dot{\phi}, \dot{\theta}, \dot{\psi}]$. 
The yaw angle $\psi$ in the global frame is excluded since footsteps are defined relative to the root frame.  

The external state $s^{\text{external}}_t \in \mathbb{R}^{10}$ includes the next two planned footsteps,  
\begin{equation}
T_i = [x_i, y_i, z_i, \theta_i], \quad i \in \{1,2\}, 
\end{equation}
expressed in the root frame, together with a two-dimensional clock signal encoding the gait phase,
\begin{equation}
\text{Clock}_t = \big[\sin(2\pi\phi/L),\; \cos(2\pi\phi/L)\big],
\end{equation}
where $\phi$ is the phase variable and $L$ is the cycle period. This continuous representation prevents the discontinuity 
that would arise if the phase were encoded by a scalar in $[0,1]$.

3) \textbf{Action space:}  
The action $a_t \in \mathbb{R}^{12}$ specifies the desired joint position targets of the robot’s 12 actuated leg joints.  
These targets are offset by a nominal half-sitting posture and converted into torque commands by a low-gain PD controller running at $1000$\,Hz.  
The high-level policy operates at $40$\,Hz, while the PD loop ensures smooth and stable torque execution.

4) \textbf{Reward function:}  
The reward encourages stable and accurate bipedal walking while discouraging unsafe or inefficient behaviors. 
It is defined as
\begin{equation}
\begin{aligned}
R_t = &\;\alpha_{\text{force}} R^{\text{force}}_t
    + \alpha_{\text{vel}} R^{\text{vel}}_t
    + \alpha_{\text{step}} R^{\text{step}}_t \\
    &+ \alpha_{\text{orient}} R^{\text{orient}}_t
    + \alpha_{\text{height}} R^{\text{height}}_t
    + \alpha_{\text{upper}} R^{\text{upper}}_t,
\end{aligned}
\end{equation}

where $\alpha_{\text{force}}{=}0.15$, $\alpha_{\text{vel}}{=}0.15$, 
$\alpha_{\text{step}}{=}0.45$, $\alpha_{\text{orient}}{=}0.05$, 
$\alpha_{\text{height}}{=}0.05$, and $\alpha_{\text{upper}}{=}0.05$.  

The foot force term regulates contact stability,  
\begin{equation}
R^{\text{force}}_t = -\,\|F^{\text{foot}}_t\|_2^2,
\end{equation}
where $F^{\text{foot}}_t$ denotes normalized ground reaction forces.  

The foot velocity term encourages smooth swing motions,  
\begin{equation}
R^{\text{vel}}_t = -\,\|v^{\text{foot}}_t\|_2^2,
\end{equation}
where $v^{\text{foot}}_t$ is the swing-foot velocity.  

The step reward promotes accurate foot placement,  
\begin{equation}
R^{\text{step}}_t = \exp\!\big(-\,\|p^{\text{foot}}_t - T_1\|^2\big),
\end{equation}
where $p^{\text{foot}}_t$ is the swing-foot position and $T_1$ is the next target.  

The orientation term aligns the root orientation with the desired target,  
\begin{equation}
R^{\text{orient}}_t = \exp\!\big(-10 \cdot (1 - \langle q_t, \hat{q}_t \rangle^2 )\big),
\end{equation}
where $q_t$ is the root quaternion and $\hat{q}_t$ is the reference.  

The body height term maintains a nominal root height,  
\begin{equation}
R^{\text{height}}_t = \exp\!\big(-40 \cdot (h^{\text{root}}_t - \hat{h}^{\text{root}})^2\big).
\end{equation}

The upper-body term reduces swaying,  
\begin{equation}
R^{\text{upper}}_t = \exp\!\big(-10 \cdot \|p^{\text{head}}_{xy} - p^{\text{root}}_{xy}\|^2\big).
\end{equation}

This compact reward balances foot-level accuracy with whole-body stability, facilitating robust locomotion learning.

\subsection{Mamba Encoder Backbone}
\label{sec:mamba-encoder}

The proposed framework employs a single-layer Mamba encoder as a lightweight backbone for processing multimodal observations. 
Unlike recurrent or transformer-based designs, this encoder operates on the current state only, avoiding temporal accumulation while still capturing structured dependencies among input features. 
The design can be described in three components:

1) \textbf{Tokenization of inputs:}  
At each time step $t$, the robot-centric state $s_t^{\text{robot}} \in \mathbb{R}^{D_r}$ and the external state 
$s_t^{\text{ext}} \in \mathbb{R}^{D_e}$ are first concatenated:
\begin{equation}
s_t \;=\; \big[\, s_t^{\text{robot}};\; s_t^{\text{ext}} \,\big] \in \mathbb{R}^{D_r + D_e}.
\end{equation}
A lightweight projection module $f_{\text{proj}}$ then maps this vector into two modality-specific embeddings:
\begin{equation}
z_t^{\text{robot}},\; z_t^{\text{ext}} \;=\; f_{\text{proj}}(s_t), 
\qquad z_t^{\text{robot}}, z_t^{\text{ext}} \in \mathbb{R}^d.
\end{equation}
The final token sequence is
\begin{equation}
U_t \;=\; \big[\, z_t^{\text{robot}};\; z_t^{\text{ext}} \,\big] \in \mathbb{R}^{2 \times d},
\end{equation}
which is passed to the Mamba backbone.

2) \textbf{State-space encoding:}  
The Mamba encoder applies gated state-space dynamics to each token:
\begin{equation}
x_{k+1} = \sigma(W_A u_{t,k})\,x_{k} + \sigma(W_B u_{t,k})\,u_{t,k},
\end{equation}
\begin{equation}
y_{t,k} = \sigma(W_C u_{t,k})\,x_{k} + \sigma(W_D u_{t,k})\,u_{t,k},
\end{equation}
where $u_{t,k}$ is the $k$-th token, $x_k$ is the latent state, and $\sigma(\cdot)$ denotes elementwise gates. 
This formulation yields a convolutional view with exponentially decaying kernels, enabling efficient feature mixing without explicit temporal recurrence.

3) \textbf{Output representation:}  
The output tokens are aggregated into a compact feature vector:
\begin{equation}
h_t = f_{\text{head}}\!\left([\,y_t^{\text{robot}};\,y_t^{\text{ext}}]\right),
\end{equation}
which serves as the shared representation for the policy and value networks.

\begin{remark}
For humanoid locomotion, the single-layer Mamba encoder provides three key advantages.  
First, it promotes energy-efficient control by producing smoother representations that reduce unnecessary torque fluctuations.  
Second, its near-linear complexity ensures computational efficiency, which is critical for real-time deployment.  
Third, the selective gating dynamics improve robustness to observation noise and enhance stability under diverse conditions.  
Together, these properties establish the Mamba encoder as an effective backbone for compact, state-centric policy learning.  
\end{remark}

\subsection{Policy Optimization with PPO}

For policy training, we adopt PPO, a widely used on-policy reinforcement learning algorithm well-suited for continuous control tasks. PPO improves stability by constraining policy updates through a clipping mechanism, thereby avoiding abrupt changes that may destabilize training \cite{schulman2017proximal}. Formally, the clipped surrogate objective is defined as
\begin{equation}
\pi^\ast = \arg\max_\pi \;
\mathbb{E}_t \Big[ \min\!\big(\rho_t(\theta) \hat{A}_t,\;
\text{clip}(\rho_t(\theta), 1-\epsilon, 1+\epsilon)\,\hat{A}_t \big)\Big],
\end{equation}
where $\pi$ is the policy parameterized by $\theta$, $\hat{A}_t$ denotes the advantage estimate at time $t$, and $\rho_t(\theta)$ is the importance sampling ratio:
\begin{equation}
\rho_t(\theta) = \frac{\pi_\theta(a_t \mid s_t)}{\pi_{\theta_{\text{old}}}(a_t \mid s_t)}.
\end{equation}
The hyperparameter $\epsilon$ specifies the clipping range, limiting the extent of each policy update.  

The advantage $\hat{A}_t$ is estimated using Generalized Advantage Estimation (GAE) to reduce variance:
\begin{equation}
\delta_t = r_t + \gamma V_\phi(s_{t+1}) - V_\phi(s_t), \qquad
\hat{A}_t = \sum_{l=0}^{T-t-1} (\gamma \lambda)^l \, \delta_{t+l},
\end{equation}
with discount factor $\gamma$ and smoothing parameter $\lambda$ controlling the bias–variance trade-off.  

The critic is optimized by minimizing the squared error between predicted values and empirical returns:
\begin{equation}
\mathcal{L}_V(\phi) = \tfrac{1}{2}\,\mathbb{E}_t \!\left[ \big(V_\phi(s_t) - \hat{R}_t\big)^2 \right],
\end{equation}
where $\hat{R}_t = \sum_{k=t}^T \gamma^{k-t} r_k$ denotes the discounted return.  

To encourage exploration, an entropy bonus is added:
\begin{equation}
\mathcal{H}_t = - \sum_a \pi_\theta(a \mid s_t)\,\log \pi_\theta(a \mid s_t).
\end{equation}

The overall training loss combines these objectives as
\begin{equation}
\mathcal{J}(\theta,\phi) = -\mathcal{L}_{\text{clip}}(\theta) 
+ \beta_V \mathcal{L}_V(\phi) 
- \beta_H \mathbb{E}[\mathcal{H}_t],
\end{equation}
where $\beta_V$ and $\beta_H$ are weighting factors for the value function and entropy terms, respectively.  
This formulation provides a stable and efficient optimization scheme for training policies based on our single-layer Mamba encoder.

\section{Experiments}
\label{sec:fourth}

\subsection{Environment Setup}
All experiments are carried out on a laptop with an 11th Gen Intel(R) Core i9-11900K CPU (16~cores, 3.5\,GHz base frequency) and a NVIDIA GeForce RTX 3060 Ti GPU (8\,GB, CUDA~12.4). The computing environment is based on Ubuntu~22.04, with physics simulation executed in \texttt{mc-mujoco} \cite{singh2023mc}. All models are developed in Python~3.12.4 using Ray~2.40.0 and PyTorch~2.4.1. All sampling and training are conducted on the GPU, while multi-thread techniques are adopted on CPU to accelerate the whole learning process in parallel as well. 

We conduct experiments on the JVRC-1 humanoid robot, a virtual platform originally designed for the Japan Virtual Robotics Challenge \cite{okugawa2015proposal}. 
The model has a height of $1.72$\,m and a mass of $62$\,kg, and is simulated with full-body dynamics. 
For locomotion tasks, only the leg joints are actuated, including hip pitch, hip roll, hip yaw, knee, ankle roll, and ankle pitch.

The proposed HuMam is trained and evaluated across multiple locomotion scenarios that vary in gait requirements and trajectory complexity:  
\begin{itemize}
  \item \textbf{Forward walking}: the robot follows a straight trajectory starting from its root projection, alternating left and right footsteps.  
  \item \textbf{Backward walking}: footsteps are mirrored along a backward line segment while maintaining a forward-facing torso.  
  \item \textbf{Curved walking}: footsteps are generated by a footstep planner to follow a curved path with varying headings.  
  \item \textbf{Standing}: a degenerate case where both target steps are set to the origin, testing balance without displacement.  
  \item \textbf{Lateral walking}: footsteps are placed along the frontal plane, enabling sideways motion with a forward torso orientation.  
\end{itemize}

\begin{figure*}[t]
\centering
% ---------- Subfigure (a) ----------
\begin{subfigure}{0.19\textwidth}
\centering
\includegraphics[clip, trim=200pt 4pt 250pt 0pt, height=90pt, width=\linewidth]
{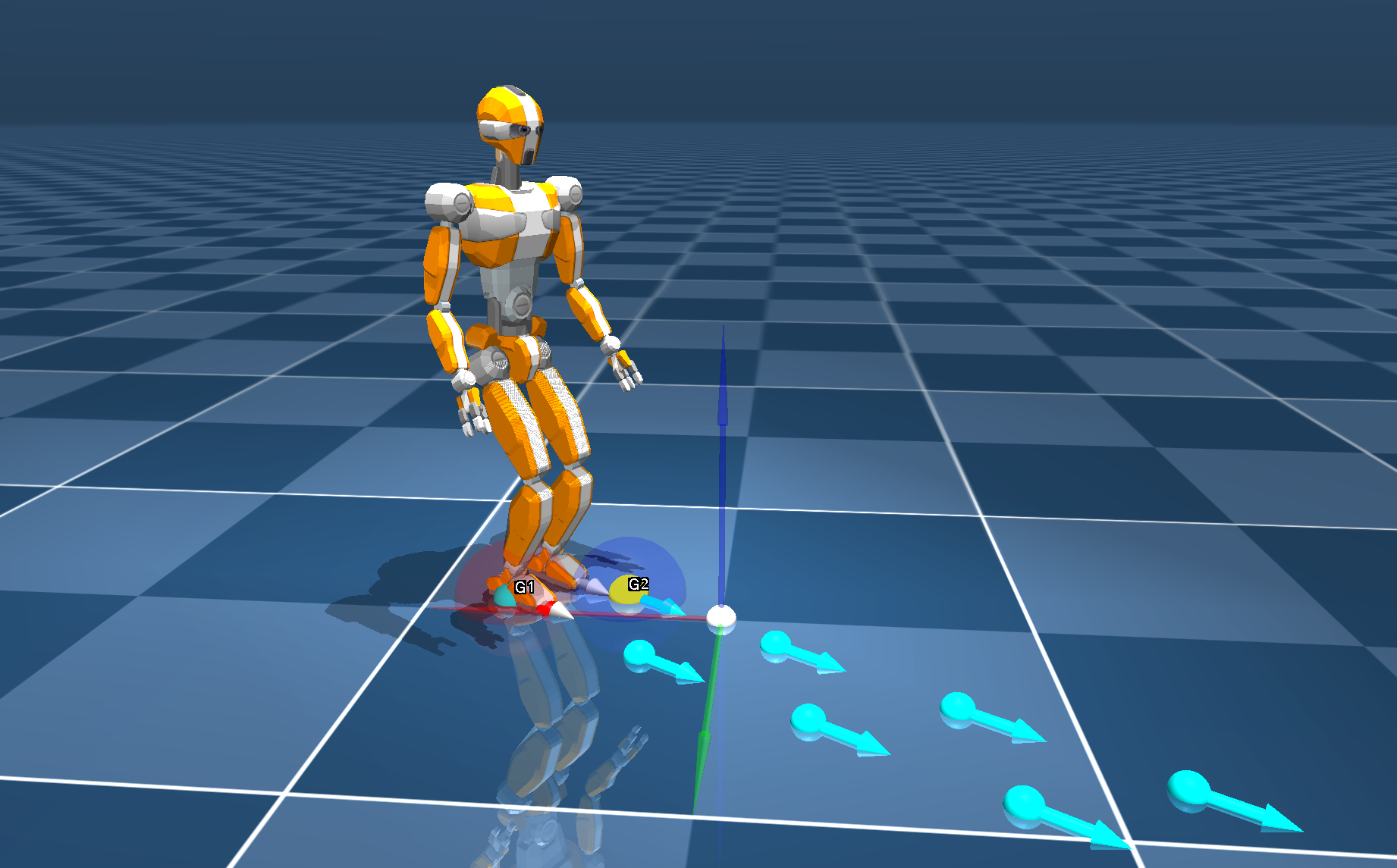}
    \caption{}
    \label{fig:suba}
\end{subfigure}
% ---------- Subfigure (b) ----------
\begin{subfigure}{0.19\textwidth}
\centering
\includegraphics[clip, trim=500pt 150pt 500pt 150pt, height=90pt, width=\linewidth]
{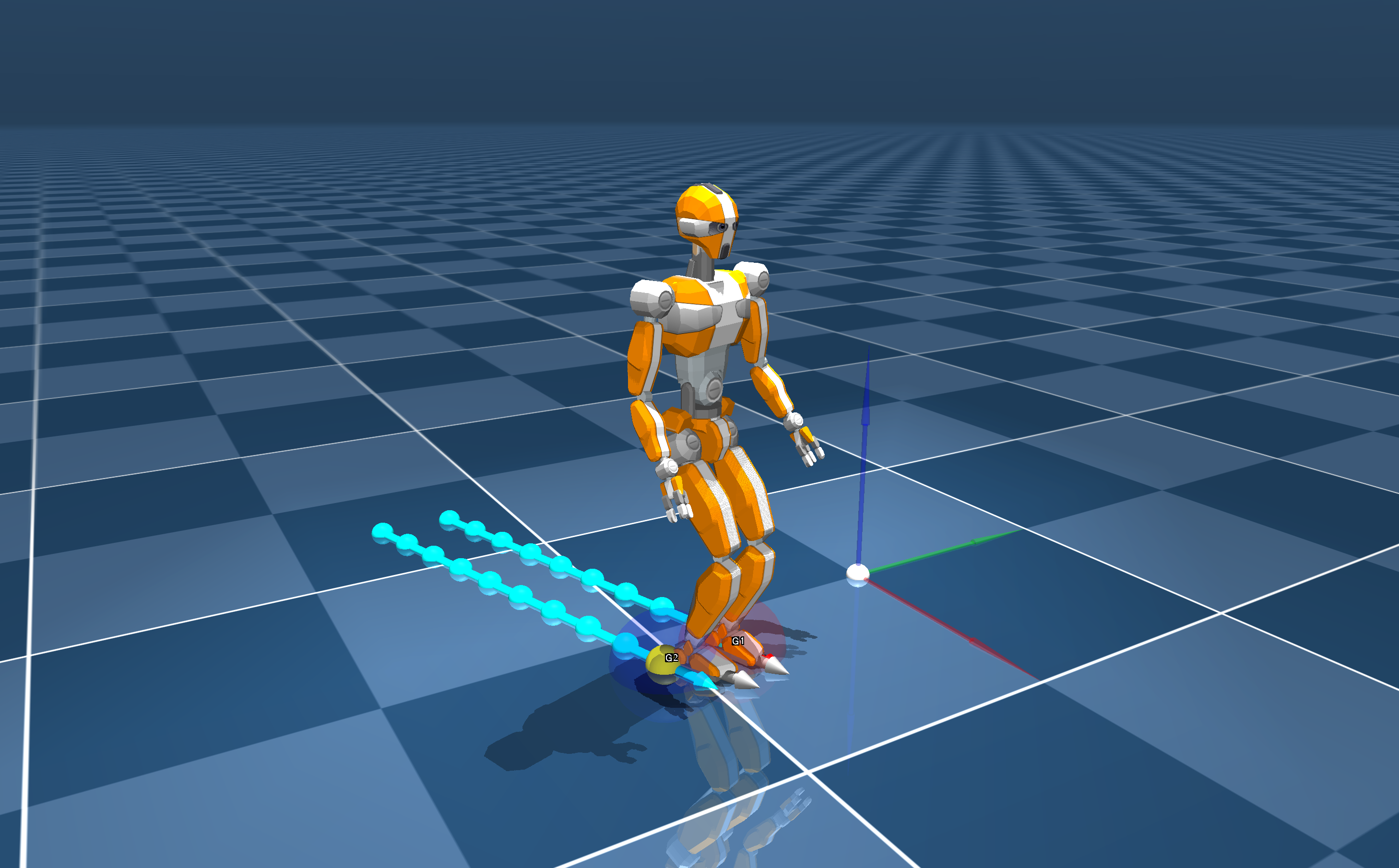}
    \caption{}
    \label{fig:subb}
\end{subfigure}
% ---------- Subfigure (c) ----------
\begin{subfigure}{0.19\textwidth}
\centering
\includegraphics[clip, trim=200pt 4pt 200pt 2pt, height=90pt, width=\linewidth]
{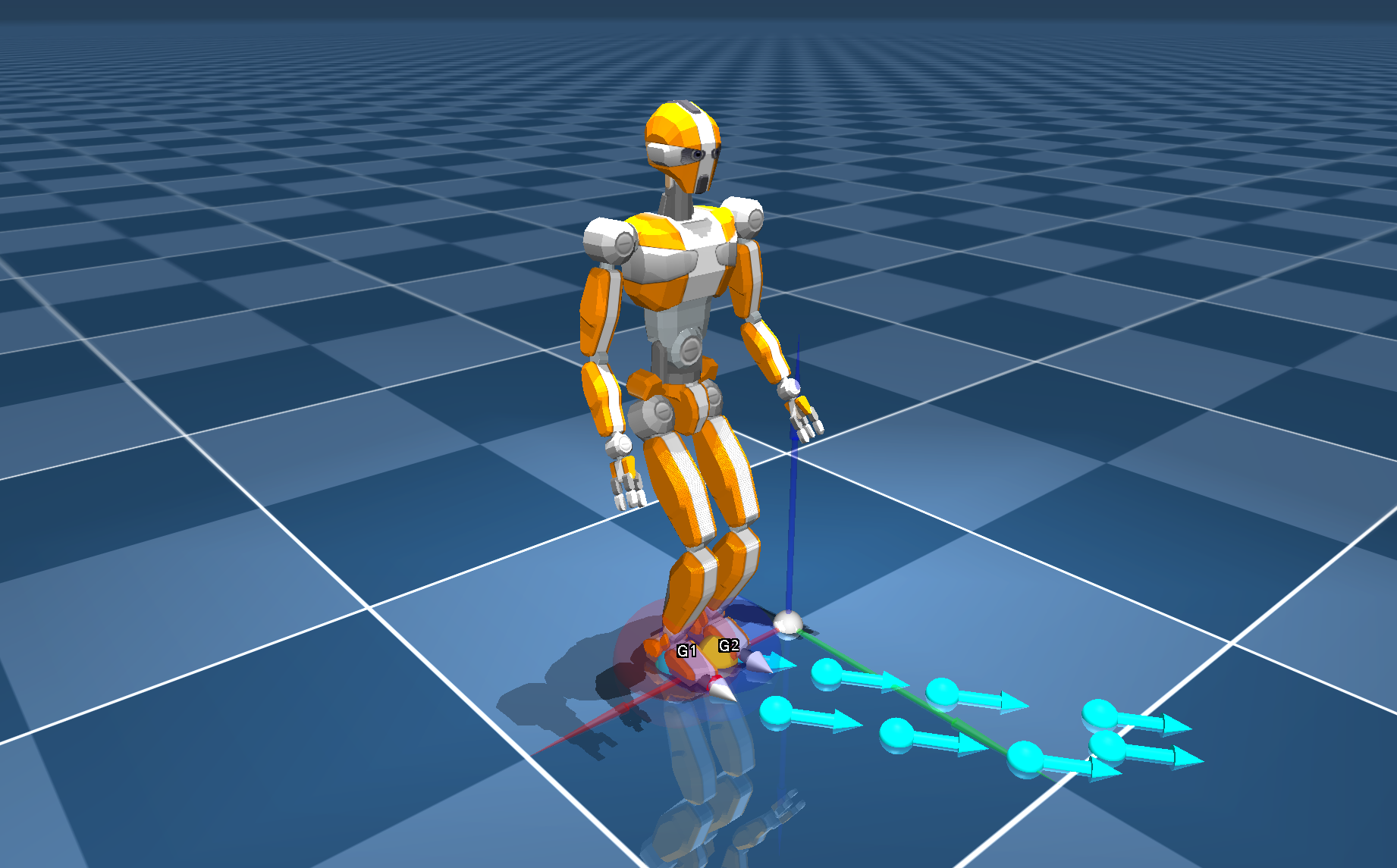} 
    \caption{}
    \label{fig:subc}
\end{subfigure}
% ---------- Subfigure (d) ----------
\begin{subfigure}{0.19\textwidth}
\centering
\includegraphics[clip, trim=300pt 50pt 400pt 40pt, height=90pt, width=\linewidth]
{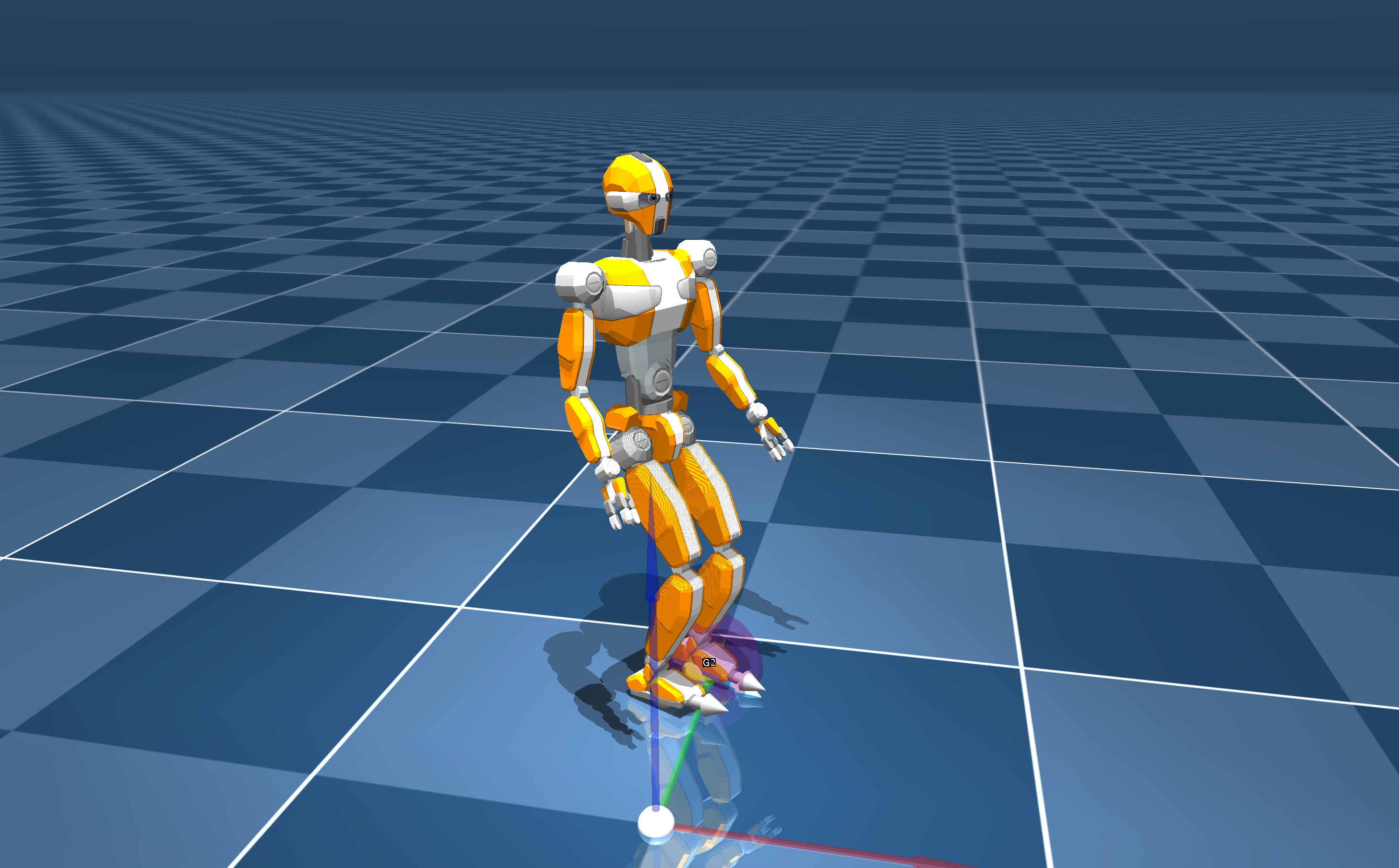}
    \caption{}
    \label{fig:subd}
\end{subfigure}
% ---------- Subfigure (e) ----------
\begin{subfigure}{0.19\textwidth}
\centering
\includegraphics[clip, trim=250pt 0pt 200pt 0pt, height=90pt, width=\linewidth]
{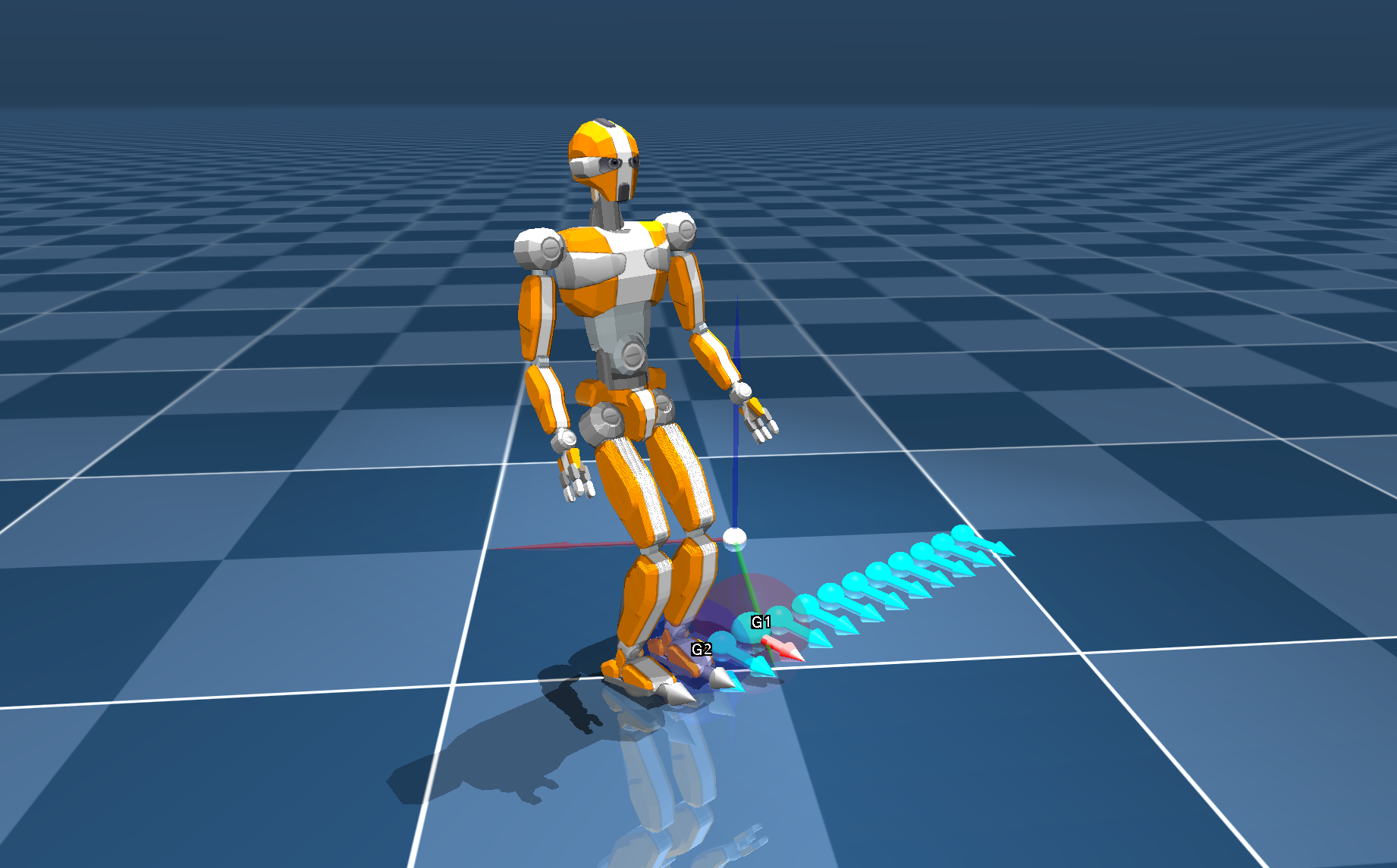}
    \caption{}
    \label{fig:sube}
\end{subfigure}
\caption{Simulated environments that the robot is trained and evaluated. Panels (a)–(e): (a) Walking straight forward; (b) Walking straight backward; (c) Walking on a curved path; (d) Standing in place; (e) Lateral Walking.}
\label{fig:env_repre}
\end{figure*}

Representative examples of these environments are shown in Fig.~\ref{fig:env_repre}. Unless specified otherwise, footstep targets are regenerated at episode resets; the robot is always provided with the next two upcoming targets relative to its root frame.

We compared our HuMam against \emph{Baseline} without using the Mamba encoder \cite{singh2022learning}. Instead, the observations are processed by a conventional feedforward network under identical training settings.

\subsection{Implementation Details}
\paragraph{Model architecture.} 
As shown in Fig.~\ref{fig:overall_archi}, the policy network consists of a single-layer Mamba encoder for feature fusion, and compact projection heads for policy and value estimation. 
This design ensures low computational overhead while retaining structured feature interactions. 
The architectural settings are summarized in Table~\ref{tab:arch}. 

\begin{table}[t]
\centering
\small
\setlength{\tabcolsep}{6pt}
\caption{Network architecture settings.}
\label{tab:arch}
\begin{tabularx}{\linewidth}{
  @{\hspace{4pt}}
  >{\hspace{2pt}}l
  >{\hspace{2pt}\raggedright\arraybackslash}X
  @{\hspace{4pt}}
}
\toprule
\textbf{Component} & \textbf{Configuration} \\
\midrule
Token width $d$ & 41 \\
Mamba encoder & Single-layer Mamba, hidden size $d=128$ \\
Projection heads & Single-layer MLP, ReLU \\
\bottomrule
\end{tabularx}
\end{table}

\paragraph{Initialization and termination.}  
At the beginning of each episode, the robot is initialized in a nominal half-sitting posture, which allows it to remain upright without external support. To improve robustness and prevent overfitting, small Gaussian noise is added to the joint positions. Since the global yaw angle of the root is not part of the observation space, the initial heading is kept fixed, while the gait phase variable $\phi$ is initialized randomly to either $0$ or $0.5$.  

Episodes are terminated early when recovery becomes infeasible. Two termination conditions are enforced: (i) a fall, defined when the root height drops below $0.60$\,m relative to the lowest foot–floor contact, and (ii) a self-collision event. Otherwise, the rollout continues until a fixed horizon $T$ is reached.  

\paragraph{Footstep generation.}  
In our framework, each footstep is defined as a 3D target location with an associated heading angle $\theta$, specifying both the desired foot placement and the torso yaw orientation. By attaching a heading to every step, the robot is able to perform behaviors such as lateral walking and turning in place. A sequence of such steps forms a footstep plan.  

For straight walking, plans are manually constructed by alternating steps to the left and right of a forward line segment projected from the root. The step length and foot spread are chosen based on classical humanoid controllers to ensure realistic gait patterns (e.g. $0.25$\,m step length and $0.12$\,m foot spread for JVRC-1). Backward walking is obtained by extending the line segment in the opposite direction, while standing is represented by a degenerate plan with $T_1=T_2=0$. Lateral walking is constructed by placing forward-facing steps along the intersection of the frontal plane and the floor.  

Curved walking plans are generated using the Humanoid Navigation ROS footstep planner\cite{hornung2012anytime}, which computes sequences on a 2D occupancy grid given start and goal poses $(x,y,\theta)$. We generate $1000$ trajectories by fixing the start at the origin and sampling goals within $(0,-1,-\pi/2)$ to $(0,1,\pi/2)$, with additional complexity introduced by randomly placed obstacles. 

During execution, the policy observes two consecutive steps relative to the root frame. Once a target at index $k$ is achieved, the observation window advances to steps $k+1$ and $k+2$. A step is considered scored if any foot remains within a $20$\,cm radius of the target for at least one single-support duration, ensuring temporal alignment between the stepping sequence and the underlying gait cycle.

\paragraph{Training schema.} 
All policies are trained with PPO using on-policy rollouts of length $T$, minibatch updates over multiple epochs, advantage normalization, and gradient clipping. The overall training procedure is outlined in Algorithm~\ref{alg:humam}, and the shared PPO hyperparameters are listed in Table~\ref{tab:rl}.

\begin{table}[t]
\centering
\caption{PPO hyperparameters.}
\label{tab:rl}
\begin{tabular*}{\linewidth}{
  @{\hspace{4pt}}
  >{\hspace{2pt}}l
  @{\extracolsep{\fill}}
  >{\hspace{2pt}}l
  @{\hspace{4pt}}
}
\toprule
\textbf{Hyperparameter} & \textbf{Value} \\
\midrule
Episode horizon (steps) & 400 \\
Samples per iteration & 4{,}800 \\
Minibatch size & 64 \\
Optimization epochs per update & 3 \\
Discount factor $\gamma$ & 0.99 \\
GAE parameter $\lambda$ & 0.95 \\
PPO clip parameter $\epsilon$ & 0.2 \\
Entropy coefficient & 0.005 \\
Policy learning rate & $1\times 10^{-4}$ \\
Value learning rate & $1\times 10^{-4}$ \\
Optimizer & Adam \\
Activation function & ReLU \\
\bottomrule
\end{tabular*}
\end{table}

\begin{algorithm}[t]
\caption{Training Procedure of HuMam with PPO and Domain Randomization}
\label{alg:humam}
\begin{algorithmic}[1]
\Require Policy parameters $\theta$, value parameters $\phi$, horizon $T$, PPO/GAE hyperparameters $(\gamma,\lambda,\epsilon,\beta_V,\beta_H)$
\Ensure Trained policy $\pi_\theta$ and value function $V_\phi$
\For{iteration $k = 1,2,\dots$}
  \State Initialize replay buffer $\mathcal{D} \gets \emptyset$
  \While{$|\mathcal{D}| < N_{\text{iter}}$}
    \State Sample domain randomization params $\xi$ (dynamics, sensors, init state, target jitter)
    \State Reset environment with $\xi$, footstep targets, and robot initialization
    \For{$t = 1$ to $T$}
      \State Observe $s_t=\{s_t^{\text{robot}},\,s_t^{\text{ext}}\}$; normalize observations
      \State Encode with single-layer Mamba: $h_t \gets f_{\text{mamba}}(s_t)$
      \State Sample action $a_t \sim \pi_\theta(\cdot \mid h_t)$
      \State Step env $\rightarrow$ obtain $(r_t, s_{t+1}, \text{done})$
      \State Store $(h_t, a_t, r_t, \log\pi_\theta(a_t \mid h_t), V_\phi(h_t), \text{done})$ in $\mathcal{D}$
      \If{done} \textbf{break} \EndIf
    \EndFor
  \EndWhile
  \State Compute returns $\hat{R}_t$ and advantages $\hat{A}_t$ with GAE; normalize $\hat{A}_t$
  \For{epoch $=1$ to $E$}
    \For{minibatch $\mathcal{B}\subset \mathcal{D}$}
      \State Compute ratio $\rho_t \gets \frac{\pi_\theta(a_t\mid h_t)}{\pi_{\theta_{\text{old}}}(a_t\mid h_t)}$
      \State Compute clipped policy loss $\mathcal{L}_{\text{clip}}$, value loss $\mathcal{L}_V$, and entropy $\mathcal{H}$
      \State Update $(\theta,\phi)$ by minimizing $-\mathcal{L}_{\text{clip}} + \beta_V \mathcal{L}_V - \beta_H \mathcal{H}$ with gradient clipping
    \EndFor
  \EndFor
\EndFor
\end{algorithmic}
\end{algorithm}

\subsection{Evaluation Metrics}
The learned policies are evaluated along four complementary dimensions:  

\begin{itemize}
    \item \textbf{Task performance.}  
  We report both peak episode returns and the mean return over the final $10\%$ of training, which together reflect the attainable performance ceiling and the consistency of long-horizon training outcomes.  

    \item \textbf{Learning efficiency and stability.}  
  Training dynamics are compared in terms of the number of environment samples required to reach specific performance thresholds (sample efficiency), the variance of learning curves, and late-stage training deviation. These measures reflect how reliably each model converges. 
  
  \item \textbf{Reward-based performance.}  
  We report the weighted per-step reward ($\in[0,0.9]$) averaged over 10,000 steps, together with a decomposition into the six reward components (foot force, foot velocity, step reward, orientation, body height, and upper-body stability). This provides a detailed view of how different terms contribute to overall locomotion quality.  

    \item \textbf{Joint torque profiles.}  
  To analyze physical efficiency, we compare average and peak torques across all actuated leg joints. Reductions in torque magnitude and variance indicate smoother, more energy-saving behaviors, and highlight the effect of structured representations in stabilizing joint actuation. 
  
  \item \textbf{Energy efficiency.}  
  To assess control economy, we measure energy cost in joules per meter (J/m), average power consumption (W), and normalized power per unit body mass (W/kg). These metrics capture both absolute and body-scaled efficiency of locomotion behaviors.

\end{itemize}

\subsection{Experiment Results}
\label{sec:results}

\subsubsection{\textbf{Performance vs.\ Baseline}}
\begin{figure}
\centering
\includegraphics[width=0.95\linewidth]{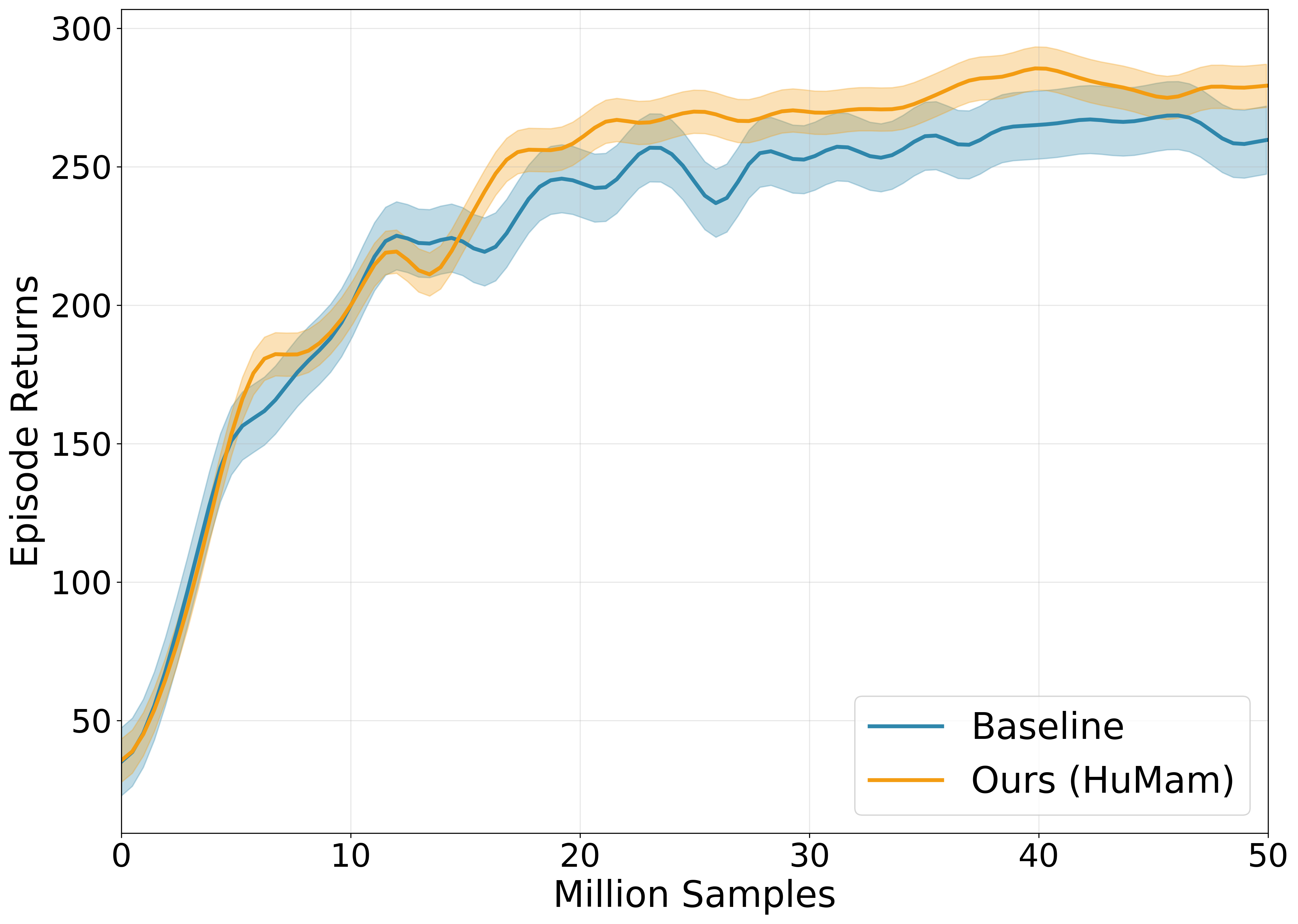}
\caption{Training curves of HuMam and Baseline across scenarios. 
Solid lines denote the mean episode return across seeds, while shaded regions indicate the standard deviation.}
\label{fig:learn_curves}
\end{figure}

\begin{figure}
\raggedleft
\includegraphics[clip, trim=250pt 1pt 1pt 1pt, width=0.4\linewidth]{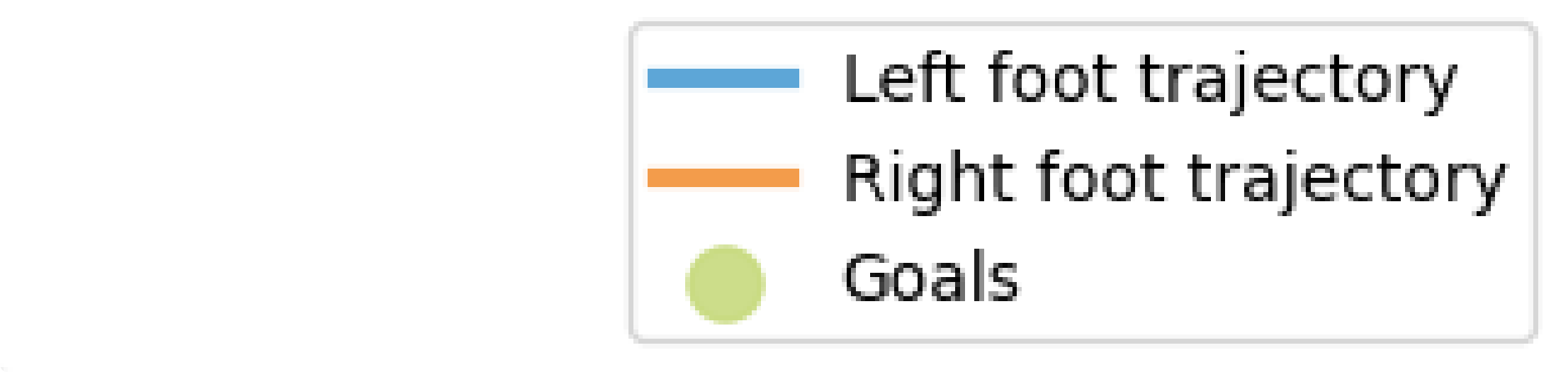}

\includegraphics[clip, trim=250pt 10pt 150pt 95pt, height=180pt, width=0.95\linewidth]{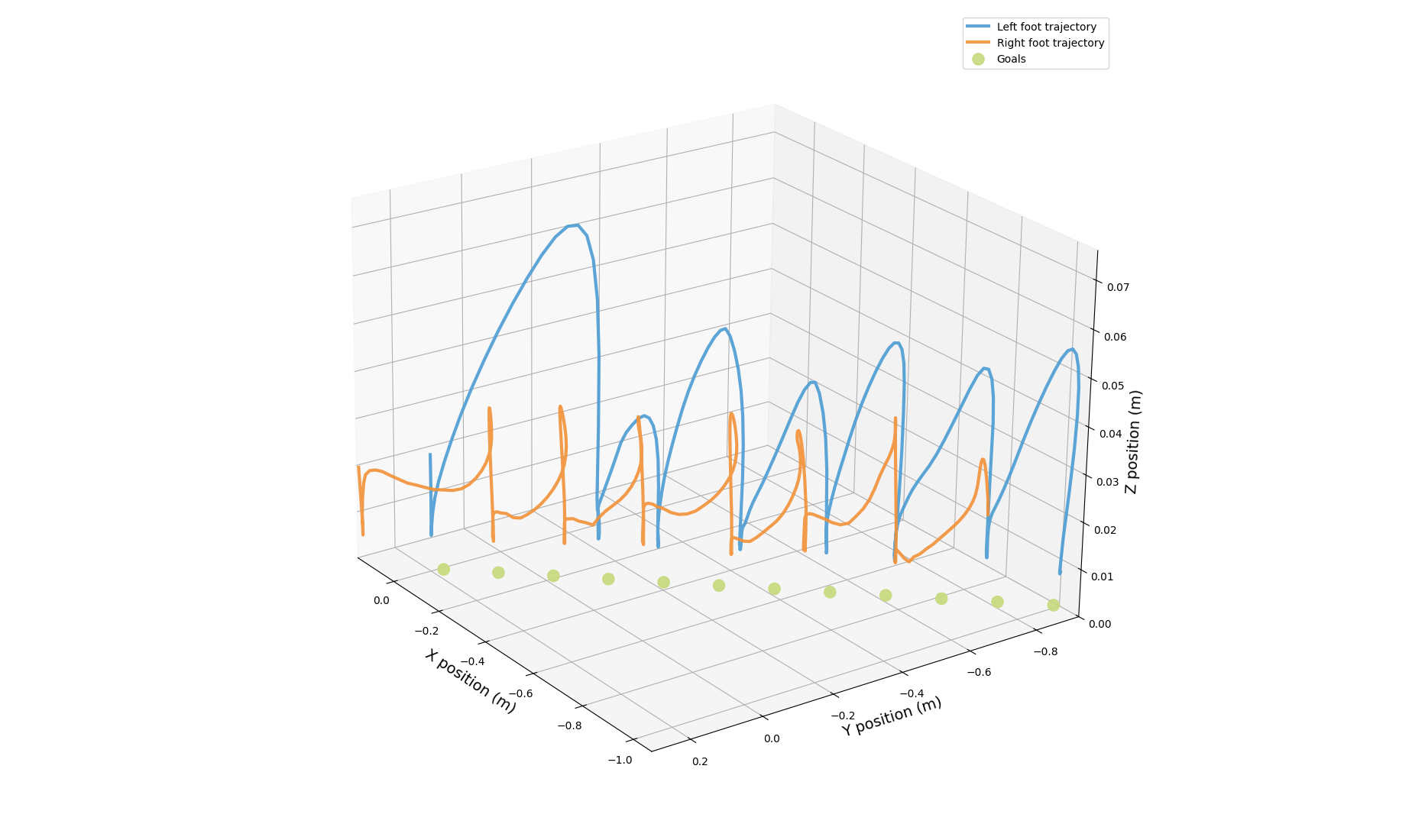}
\caption{Foot trajectory of lateral walking.}
\label{fig:lateral_foot}
\end{figure}

\begin{figure}
\raggedleft
\includegraphics[clip, trim=250pt 1pt 1pt 1pt, width=0.4\linewidth]{experiment_results/legend.png}
\includegraphics[clip, trim=250pt 10pt 150pt 95pt, height=180pt, width=0.95\linewidth]{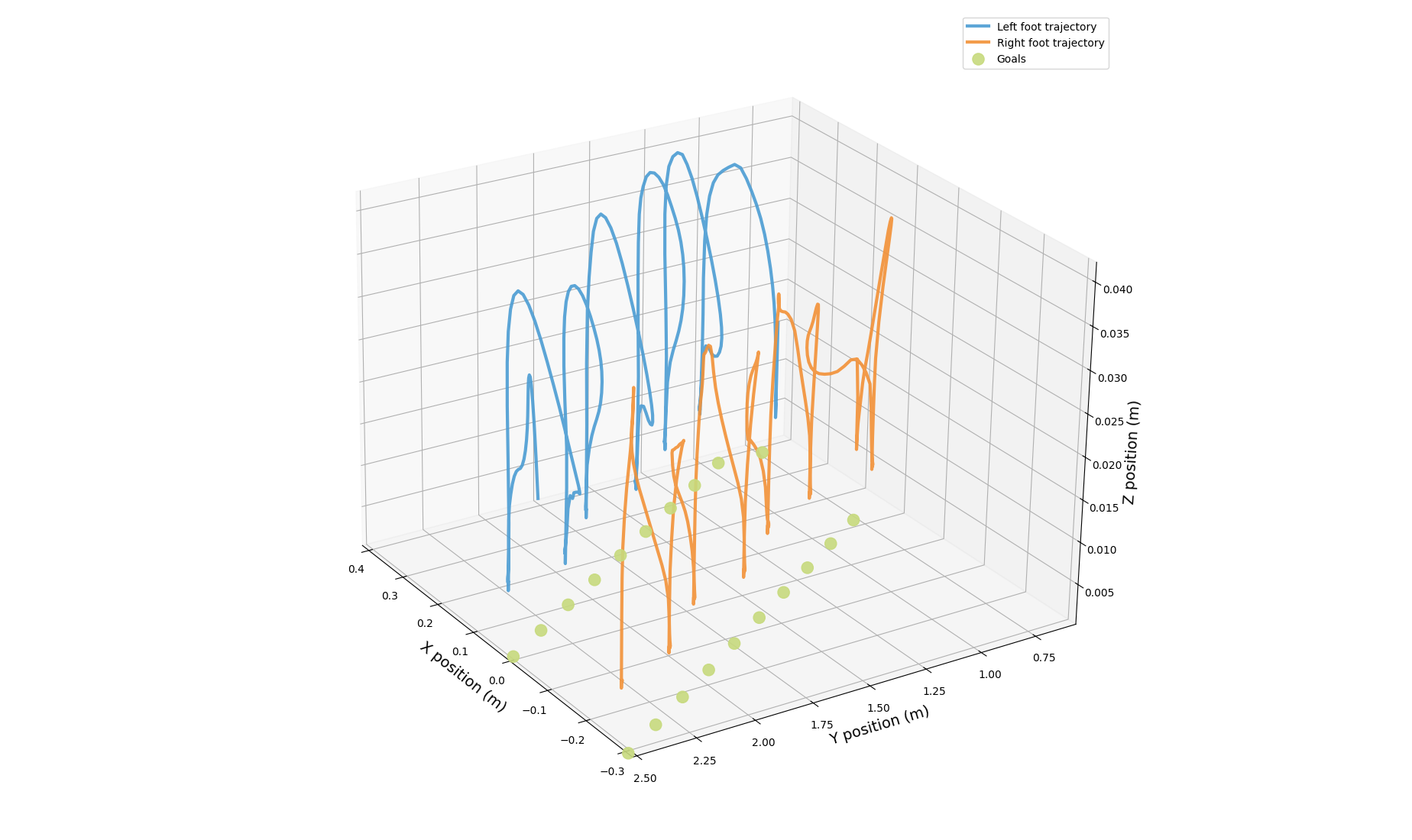}
\caption{Foot trajectory of backward walking.}
\label{fig:back_foot}
\end{figure}

\begin{figure}
\raggedleft
\includegraphics[clip, trim=250pt 1pt 1pt 1pt, width=0.4\linewidth]{experiment_results/legend.png}
\includegraphics[clip, trim=250pt 10pt 150pt 95pt, height=180pt, width=0.95\linewidth]{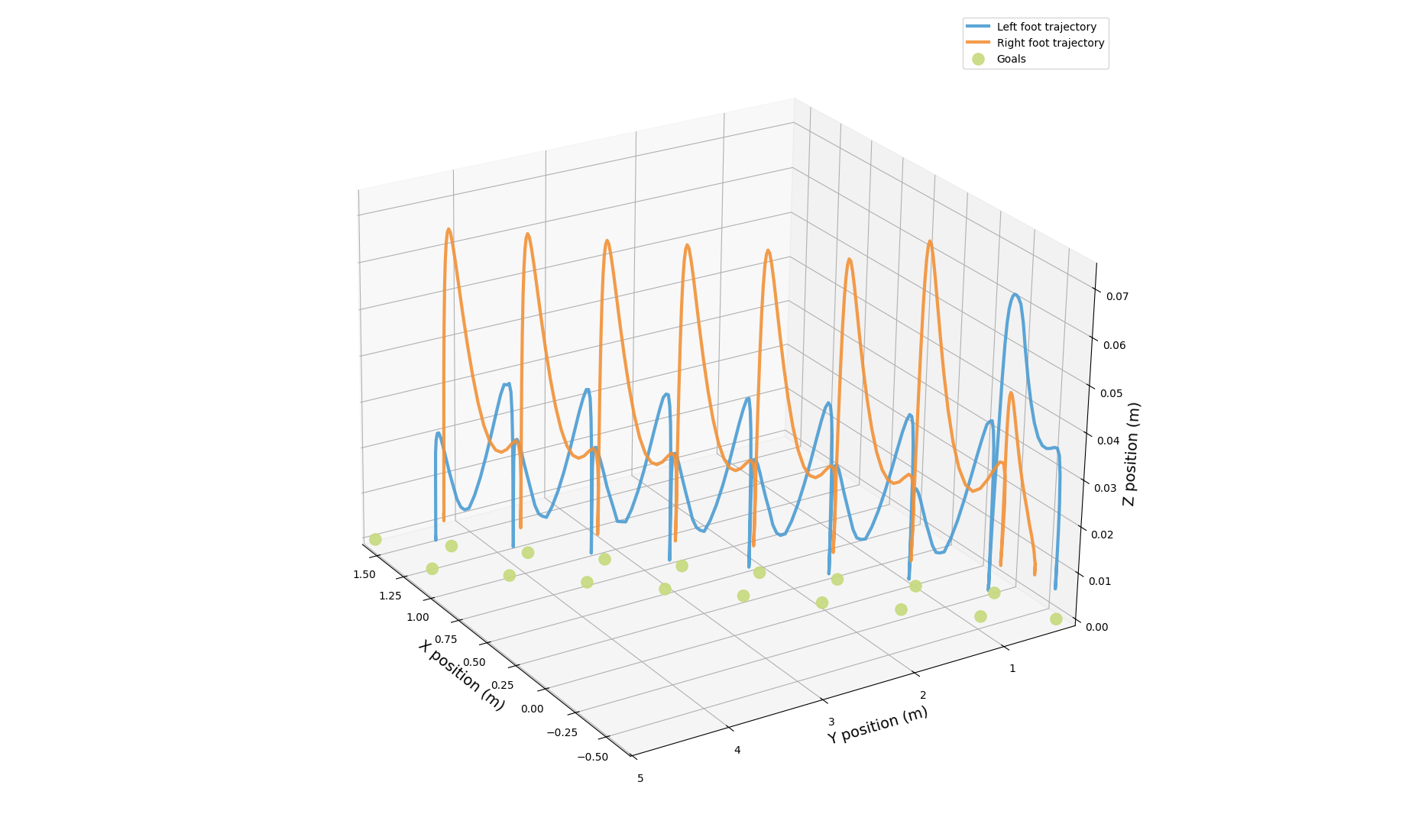}
\caption{Foot trajectory of forward walking.}
\label{fig:for_foot}
\end{figure}

\begin{figure}
\raggedleft
\includegraphics[clip, trim=250pt 1pt 1pt 1pt, width=0.4\linewidth]{experiment_results/legend.png}
\includegraphics[clip, trim=180pt 10pt 150pt 65pt, height=180pt, width=0.95\linewidth]{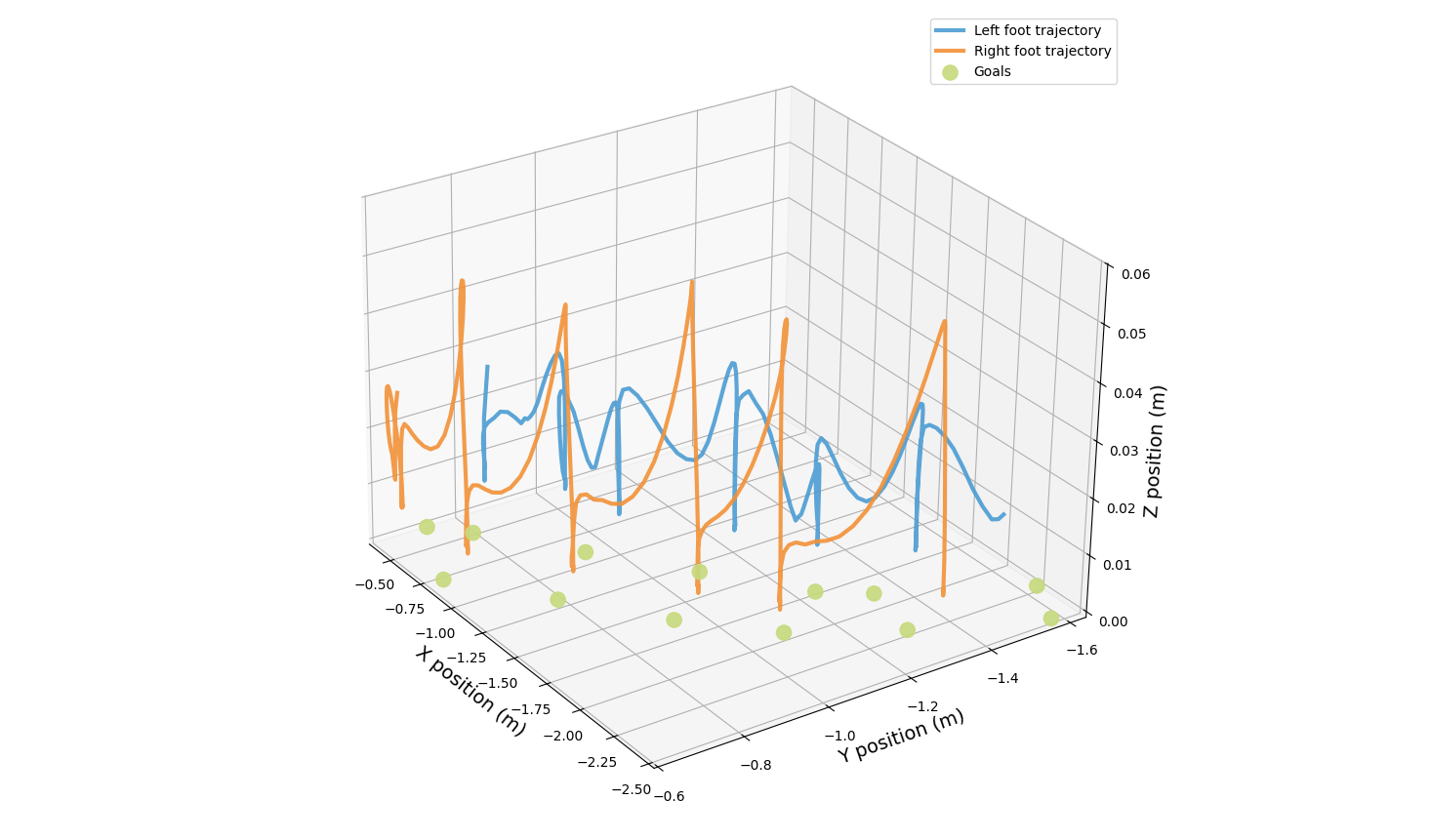}
\caption{Foot trajectory of curved path walking.}
\label{fig:curve_foot}
\end{figure}

Fig.~\ref{fig:learn_curves} compares the learning curves of HuMam and the Baseline across seeds, 
where solid lines denote the mean return and shaded regions indicate one standard deviation.  
HuMam consistently achieves higher returns, with a steeper rise at the beginning and a higher plateau in the later stage.  Fig. \ref{fig:lateral_foot}, \ref{fig:back_foot}, 
\ref{fig:for_foot}, and \ref{fig:curve_foot} show the foot trajectories for lateral, backward, forward, and curved walking, respectively.
Table~\ref{tab:performance_comparison} further quantifies these differences: 
HuMam improves peak performance from $269.85$ to $285.50$ (+\textbf{5.8}\%) 
and raises the mean return over the last $10\%$ of training from $263.03$ to $277.50$ (+\textbf{5.5}\%).  
Together, these results demonstrate that HuMam not only achieves a higher performance ceiling but also sustains stronger outcomes in long-horizon training.

\begin{table*}[t]
\centering
\begin{minipage}{0.75\linewidth}
  \centering
  \caption{Comparison of Learning Performance. Mean $\pm$ std over multiple runs.}
  \label{tab:performance_comparison}
  \small
  \begin{tabular*}{\linewidth}{
    @{\hspace{4pt}}
    l
    @{\extracolsep{\fill}}
    c c c
    @{\hspace{4pt}}
  }
  \toprule
  \textbf{Metric} & \textbf{Baseline} & \textbf{Ours} & \textbf{Improvement} \\
  \midrule
  \multicolumn{4}{l}{\textit{Sample Efficiency}} \\
  Samples to reach 240 returns (M)  & 18.2  & \textbf{15.8} & 13.2\% \\
  Samples to reach 260 returns (M)  & 35.0  & \textbf{20.2} & 42.5\% \\
  \midrule
  \multicolumn{4}{l}{\textit{Learning Stability}} \\
  Learning curve deviation          & 12.08 & \textbf{7.81} & 35.4\% \\
  Late training variance            & 102.10 & \textbf{39.87} & 61.0\% \\
  \midrule
  \multicolumn{4}{l}{\textit{Final Performance}} \\
  Peak performance                  & 269.85 & \textbf{285.50} & 5.8\% \\
  Final performance (last 10\%)     & 263.03 & \textbf{277.50} & 5.5\% \\
  \bottomrule
  \end{tabular*}
\end{minipage}
\end{table*}

\subsubsection{\textbf{Effect on Learning Efficiency and Stability}}
As shown in Fig.~\ref{fig:learn_curves}, HuMam exhibits a steeper early slope, higher asymptotic return, and narrower confidence intervals, suggesting faster and more consistent learning.  

Quantitative results in Table~\ref{tab:performance_comparison} further confirm these observations.  
In terms of efficiency, HuMam requires only $15.8$M samples to reach a return of $240$, compared to $18.2$M for the Baseline (\textbf{13.2\%} fewer samples).  
The gap widens at higher thresholds, with HuMam achieving $260$ returns in $20.2$M samples versus $35.0$M for the Baseline, a \textbf{42.5\%} improvement.  

For stability, HuMam reduces cross-seed learning curve deviation from $12.08$ to $7.81$ (\textbf{35.4\%} lower) and late-stage variance from $102.10$ to $39.87$ (\textbf{61.0\%} lower).  
These reductions highlight that HuMam not only accelerates convergence but also delivers more reliable and robust training dynamics across seeds.

\subsubsection{\textbf{Effect on Reward-Based Performance}}
Table~\ref{tab:reward_comparison} provides a detailed breakdown of the six reward components across tasks.  
Overall, HuMam achieves higher or comparable scores in nearly all dimensions, leading to an average per-step reward of $0.737$, surpassing the Baseline ($0.728$).  

In particular, HuMam consistently improves \emph{foot force regulation} (0.132 vs.\ 0.128), reflecting smoother and more energy-saving contacts with the ground, and enhances \emph{velocity smoothness} (0.129 vs.\ 0.124), indicating better control of swing leg dynamics.  
Upper-body stability and body height regulation are also slightly improved, contributing to more balanced whole-body motion.  
These results align with our design goal of promoting \emph{energy-efficient and stable locomotion}.  

Although Baseline shows marginally higher step-placement accuracy in curved and forward walking, HuMam achieves better balance across all reward terms.  
This indicates that the Mamba encoder does not over-optimize a single component (e.g., foot placement) at the expense of global stability, but instead integrates multimodal inputs into a more \emph{holistic control strategy}.  

By embedding structured dependencies among robot-centric and external states through selective gating, HuMam learns to reduce unnecessary torque usage, stabilize contacts, and improve overall coordination.  
This validates our claim that the Mamba encoder contributes to \emph{efficiency, stability, and energy saving}, delivering more reliable locomotion than conventional feedforward representations.

\begin{table*}[t]
\centering
\caption{Comprehensive Reward Components Comparison}
\label{tab:reward_comparison}
\resizebox{\linewidth}{!}{
\begin{tabular}{@{}lcccccccccccccc@{}}
\toprule
\multirow{2}{*}{\textbf{Task}} & \multicolumn{2}{c}{\textbf{Foot Force}} & \multicolumn{2}{c}{\textbf{Foot Velocity}} & 
\multicolumn{2}{c}{\textbf{Foot Step}} & \multicolumn{2}{c}{\textbf{Orientation}} & \multicolumn{2}{c}{\textbf{Body Height}} & 
\multicolumn{2}{c}{\textbf{Upper Body}} & \multicolumn{2}{c}{\textbf{Total}}\\
\cmidrule(lr){2-3} \cmidrule(lr){4-5} \cmidrule(lr){6-7} \cmidrule(lr){8-9} \cmidrule(lr){10-11} \cmidrule(lr){12-13} \cmidrule(lr){14-15}
& \textbf{Ours} & \textbf{Baseline} & \textbf{Ours} & \textbf{Baseline} & \textbf{Ours} & \textbf{Baseline} & \textbf{Ours} & \textbf{Baseline} & \textbf{Ours} & \textbf{Baseline} & \textbf{Ours} & \textbf{Baseline} & \textbf{Ours} & \textbf{Baseline} \\
\midrule
Standing & \textbf{0.131} & 0.102 & \textbf{0.143} & 0.134 & \textbf{0.423} & 0.413 & \textbf{0.049} & 0.048 & \textbf{0.041} & 0.039 & \textbf{0.049} & 0.048  & \textbf{0.837} & 0.785 \\
Lateral  & \textbf{0.133} & \textbf{0.133} & \textbf{0.124} & 0.123 & \textbf{0.351} & \textbf{0.351} & \textbf{0.049} & 0.047 & \textbf{0.045} & 0.043 & \textbf{0.049} & 0.048 & \textbf{0.752} & 0.746 \\
Curved   & 0.132 & \textbf{0.136} & \textbf{0.125} & 0.122 & 0.279 & \textbf{0.284} & 0.042 & \textbf{0.044} & \textbf{0.044} & 0.042 & \textbf{0.049} & 0.048 & 0.672 & \textbf{0.677} \\
Forward  & 0.132 & \textbf{0.139} & \textbf{0.128} & 0.121 & 0.270 & \textbf{0.299} & \textbf{0.049} & 0.047 & \textbf{0.041} & \textbf{0.041} & \textbf{0.049} & 0.047 & 0.668 & \textbf{0.694} \\
Backward & \textbf{0.131} & 0.129 & \textbf{0.127} & 0.121 & \textbf{0.352} & 0.349 & \textbf{0.049} & 0.048 & \textbf{0.046} & 0.044 & \textbf{0.049} & \textbf{0.049} & \textbf{0.754} & 0.740 \\
\midrule
Average & \textbf{0.132} & 0.128 & \textbf{0.129} & 0.124 & 0.335 & \textbf{0.339} & \textbf{0.048} & 0.047 & \textbf{0.043} & 0.042 & \textbf{0.049} & 0.048 & \textbf{0.737} & 0.728 \\
\bottomrule
\end{tabular}
}
\end{table*}

\subsubsection{\textbf{Effect on Joint Torque Profiles}}
\begin{figure*}
\centering
% \includegraphics[height=100pt, width=0.24\linewidth]{images/forward/screenshot/frame_0000000.png}
% \includegraphics[height=100pt, width=0.24\linewidth]{images/forward/screenshot/frame_0000002.png}
% \includegraphics[height=100pt, width=0.24\linewidth]{images/forward/screenshot/frame_0000004.png}
% \includegraphics[height=100pt, width=0.24\linewidth]{images/forward/screenshot/frame_0000007.png}
% Forward Walk Task\\
\includegraphics[width=\linewidth]{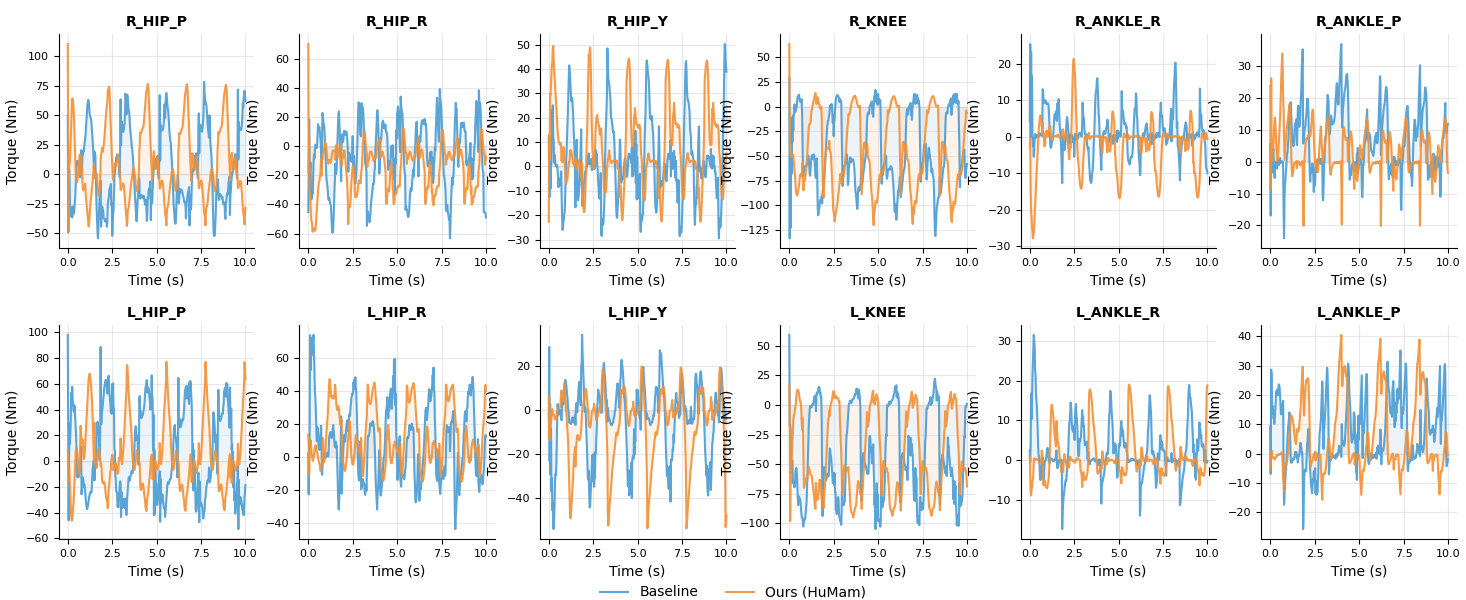}
\caption{Joint Torques of Forward Walking Task.}
\label{fig:torque}
\end{figure*}

Table~\ref{tab:joint_torque_comparison} and Fig. ~\ref{fig:torque} present the torque statistics for the forward walking task.  
On average, \emph{HuMam} reduces joint torques by \textbf{9.6\%} in magnitude and \textbf{9.1\%} in peak load,  
indicating smoother and less strenuous actuation compared with the Baseline.  

Several joints benefit most from the structured representation of the Mamba encoder.  
For instance, the left hip pitch (L\_HIP\_P) shows a substantial reduction in both average ($-37.6\%$) and peak ($-21.2\%$) torque,  
while the right ankle pitch (R\_ANKLE\_P) decreases by $-40.2\%$ in average torque.  
Similarly, stabilizing effects are observed in the knees and ankle roll joints, which are traditionally prone to high-variance actuation.  
Although a few joints (e.g., L\_KNEE, L\_ANKLE\_P) report slightly higher peak torques, these deviations are small and do not offset the overall efficiency gain.  

Taken together, the results confirm that HuMam produces more balanced torque distribution,  
reduces excessive peaks that may risk instability,  
and improves control economy across the lower-limb joints most critical for locomotion.  
This complements the improvements in reward shaping and energy efficiency,  
highlighting that the Mamba backbone not only improves learning stability but also translates into tangible physical efficiency.  

\begin{table*}[htbp]
\centering
\caption{Joint Torque Comparison of Forward Walking Task}
\label{tab:joint_torque_comparison}
\renewcommand{\arraystretch}{1.2}
\begin{tabular*}{\textwidth}{@{\extracolsep{\fill}}lcccccccc}
\hline
\toprule
\multirow{2}{*}{\textbf{Joint}}
& \multicolumn{2}{c}{\textbf{Average Torque (Nm)}}
& \multicolumn{2}{c}{\textbf{Peak Torque (Nm)}}
& \multicolumn{2}{c}{\textbf{Change \%}} \\
\cmidrule(lr){2-3} \cmidrule(lr){4-5} \cmidrule(lr){6-7} &
\textbf{Baseline} & \textbf{Ours (HuMam)} & \textbf{Baseline} & \textbf{Ours (HuMam)} & \textbf{Average Torque} & \textbf{Peak Torque} \\
& \textbf{Mean $\pm$ Std} & \textbf{Mean $\pm$ Std} & & \\
\hline
R\_HIP\_P & 27.17 $\pm$ 17.76 & \textbf{22.70 $\pm$ 20.48} & \textbf{78.28} & 110.45 & -16.4\% & 41.1\% \\
R\_HIP\_R & 19.27 $\pm$ 13.71 & \textbf{15.57 $\pm$ 13.45} & 70.45 & \textbf{70.41} & -19.2\% & -0.1\% \\
R\_HIP\_Y & \textbf{12.10 $\pm$ 11.07} & 12.30 $\pm$ 11.98 & 50.19 & \textbf{49.38} & 1.7\% & -1.6\% \\
R\_KNEE & 47.26 $\pm$ 34.10 & \textbf{47.11 $\pm$ 34.83} & 133.29 & \textbf{121.18} & -0.3\% & -9.1\% \\
R\_ANKLE\_R & 3.77 $\pm$ 4.19 & \textbf{3.77 $\pm$ 5.49} & \textbf{25.37} & 27.81 & -0.0\% & 9.6\% \\
R\_ANKLE\_P & 8.68 $\pm$ 7.61 & \textbf{5.19 $\pm$ 5.13} & 36.84 & \textbf{33.90} & -40.2\% & -8.0\% \\
L\_HIP\_P & 29.41 $\pm$ 16.76 & \textbf{18.36 $\pm$ 18.28} & 98.00 & \textbf{77.19} & -37.6\% & -21.2\% \\
L\_HIP\_R & 19.42 $\pm$ 13.71 & \textbf{16.77 $\pm$ 13.96} & 74.07 & \textbf{47.22} & -13.6\% & -36.2\% \\
L\_HIP\_Y & \textbf{11.69 $\pm$ 10.74} & 11.78 $\pm$ 12.66 & 54.19 & \textbf{53.89} & 0.8\% & -0.5\% \\
L\_KNEE & \textbf{44.25 $\pm$ 32.40} & 46.77 $\pm$ 31.27 & 105.04 & \textbf{98.27} & 5.7\% & -6.4\% \\
L\_ANKLE\_R & 4.08 $\pm$ 5.05 & \textbf{3.62 $\pm$ 4.79} & 31.58 & \textbf{19.04} & -11.1\% & -39.7\% \\
L\_ANKLE\_P & \textbf{9.50 $\pm$ 8.82} & 9.87 $\pm$ 10.60 & \textbf{35.19} & 40.50 & 3.9\% & 15.1\% \\
\hline
Total & 19.72 $\pm$ 14.66 & \textbf{17.82 $\pm$ 15.24} & 133.29 & \textbf{121.18} & -9.6\% & -9.1\% \\
\bottomrule
\end{tabular*}
\begin{tablenotes}
\small
\item Note: 
Mean $\pm$ Std represents mean absolute torque $\pm$ standard deviation. Negative percentage indicates HuMam uses less torque (more efficient).
\end{tablenotes}
\end{table*}

\subsubsection{\textbf{Effect on Energy Efficiency}}

Table~\ref{tab:energy_efficiency} reports the energy efficiency, average power, and normalized power per unit mass across tasks.  
Overall, HuMam demonstrates lower power consumption and improved economy of movement compared with the Baseline.  

In standing, HuMam reduces the average power draw from $57.48$\,W to $35.08$\,W, a \textbf{39\%} decrease, corresponding to a drop in power-per-mass from $0.92$\,W/kg to $0.56$\,W/kg.  
For dynamic locomotion, HuMam consistently lowers energy cost: lateral walking requires only $963$\,J/m compared to $1395$\,J/m, and forward walking reduces cost from $673$\,J/m to $421$\,J/m.  
Backward walking also shows improvement ($853$ vs.\ $952$\,J/m), while curved walking remains competitive, with HuMam consuming slightly more energy ($1303$ vs.\ $1055$\,J/m) but compensating with smoother dynamics and reduced power-per-mass ($1.53$ vs.\ $2.11$\,W/kg).  

These results indicate that the Mamba encoder helps produce smoother joint actuation and avoids unnecessary corrective torques, directly lowering the energy required per unit distance.  
Combined with the improvements in reward composition (Section~\ref{sec:results}) and reduced joint torque magnitudes (Table~\ref{tab:joint_torque_comparison}), the evidence consistently supports that HuMam enables more \emph{energy-efficient, stable, and sustainable locomotion} than conventional feedforward baselines.

\begin{table*}[t]
\centering
\begin{minipage}{\linewidth}
  \centering
  \caption{Energy Efficiency and Power Consumption}
  \label{tab:energy_efficiency}
  \small
  \begin{tabular*}{\linewidth}{
    @{\hspace{4pt}}
    l
    @{\extracolsep{\fill}}
    c c
    c c
    c c
    @{\hspace{4pt}}
  }
  \toprule
  \multirow{2}{*}{\textbf{Task}} 
    & \multicolumn{2}{c}{\textbf{Energy Efficiency (J/m)}} 
    & \multicolumn{2}{c}{\textbf{Average Power (W)}} 
    & \multicolumn{2}{c}{\textbf{Power Efficiency (W/kg)}} \\
  \cmidrule(lr){2-3}
  \cmidrule(lr){4-5}
  \cmidrule(lr){6-7}
  & \textbf{Ours} & \textbf{Baseline}
  & \textbf{Ours} & \textbf{Baseline}
  & \textbf{Ours} & \textbf{Baseline} \\
  \midrule

  Standing  & --         & --        & \textbf{35.08}  & 57.48  & \textbf{0.56} & 0.92 \\
  Lateral   & \textbf{963}    & 1395      & \textbf{95.45}  & 121.10 & \textbf{1.53} & 1.94 \\
  Curved    & 1303      & \textbf{1055} & \textbf{95.17}  & 131.93 & \textbf{1.53} & 2.11 \\
  Forward   & \textbf{421}    & 673       & \textbf{107.59} & 183.32 & \textbf{1.72} & 2.94 \\
  Backward  & \textbf{853}    & 952       & \textbf{97.47}  & 105.46 & \textbf{1.56} & 1.69 \\
  \midrule
  Average   & \textbf{885}    & 1019      & \textbf{86.15}  & 119.86 & \textbf{1.38} & 1.92 \\
  
  \bottomrule
  \end{tabular*}
\end{minipage}
\end{table*}

\section{Conclusion}
\label{sec:fifth}

This work presented \emph{HuMam}, an end-to-end reinforcement learning framework for humanoid motion control that employs a \emph{single-layer Mamba} encoder as a compact state-centric backbone. By tokenizing robot-centric and external stepping states and processing them with selective state-space dynamics, HuMam enables efficient feature interaction without temporal accumulation. Coupled with a carefully weighted reward that balances foot–ground interaction, foot placement accuracy, body posture, and height regulation, the framework enables stable PPO training under real-time control constraints.

Across forward, backward, lateral, curved walking, and standing, HuMam consistently improves over a strong feedforward \emph{Baseline}: it achieves higher final and peak returns, reaches performance thresholds with fewer samples, and exhibits lower cross-seed variance. The learned policies demonstrate better physical economy—lower energy per meter, reduced average power, and decreased power-per-mass—and produce smoother actuation, as evidenced by reductions in average and peak joint torques. Together, these results indicate that the Mamba backbone enhances \emph{efficiency}, \emph{stability}, and \emph{energy saving} for humanoid locomotion, validating our claim that Mamba is an effective encoder for compact, end-to-end humanoid control.

Future work will focus on hardware deployment and sim-to-real transfer for JVRC-class platforms, including latency profiling and safety monitoring. We also plan to extend the observation space with onboard perception (e.g., vision or event cues) while retaining the state-centric interface, integrate online footstep planning with feasibility checks, and study multi-contact behaviors (e.g., stairs and uneven terrain) under tighter power budgets. Broader directions include comparing Mamba with alternative lightweight backbones under identical compute envelopes and incorporating safety constraints to further improve reliability in the wild.

\printcredits

\section*{Declaration of competing interest}
The authors declare that they have no known competing financial interests or personal relationships that could have appeared to influence the work reported in this paper.

\bibliographystyle{model1-num-names}

\bibliography{cas-refs}


\end{document}